\documentclass{article}
\usepackage{spconf,amsmath,graphicx}
\usepackage{algorithm}
\usepackage{algorithmic}
\usepackage{amssymb}
\usepackage{amsmath} % 数式用パッケージ
\usepackage{mathtools} % 数式表示用パッケージ
\usepackage{array}
\usepackage{longtable}
\usepackage{colortab}
\usepackage{colortbl}
\usepackage{arydshln}
\usepackage{color}
\usepackage{bm}
\usepackage[accsupp]{axessibility} 
\title{Counting Network for Learning from Majority Label}
\name{Kaito Shiku$^{1}$, Shinnosuke Matsuo$^{1}$, Daiki Suehiro$^{2}$, Ryoma Bise$^{1}$\thanks{This work was supported by JSPS KAKENHI Grant Number JP23K18509, and JST, ACT-X Grant Number JPMJAX200G, Japan.}}
\address{$^{1}$ Kyushu University, Fukuoka, Japan \\ $^{2}$ Yokohama City University, Kanagawa, Japan}
\begin{document}
\ninept
\maketitle
\begin{abstract}
%In this paper, we propose a novel machine-learning problem in multi-class MIL called learning from the majority label (LML). 
The paper proposes a novel problem in multi-class Multiple-Instance Learning (MIL) called Learning from the Majority Label (LML). In LML, the majority class of instances in a bag is assigned as the bag's label. LML aims to classify instances using bag-level majority classes. 
This problem is valuable in various applications.
Existing MIL methods are unsuitable for LML due to aggregating confidences, which may lead to inconsistency between the bag-level label and the label obtained by counting the number of instances for each class. This may lead to incorrect instance-level classification.
We propose a counting network trained to produce the bag-level majority labels estimated by counting the number of instances for each class.
This led to the consistency of the majority class between the network outputs and one obtained by counting the number of instances.
Experimental results show that our counting network outperforms conventional MIL methods on four datasets\footnote{The code is publicly available at https://github.com/Shiku-Kaito/Counting-Network-for-Learning-from-Majority-Label.}.
%Ablation studies further confirm the counting network superiority. 
\end{abstract}
\begin{keywords}
Majority Label, Counting Network, MIL
\end{keywords}

\section{Introduction}
Many methods of multiple-instance learning (MIL) have been proposed.
MIL aims to train a classification model using bag-level class labels, in which a bag consists of a set of instances. In a conventional setup of binary classification, a negative bag only contains negative instances, and a positive bag contains at least one positive instance. 
MIL is used in various fields~\cite{Oquab2015,kolesnikov2016seed,huang2018dsrg,Wang_2020_CVPR_SEAM}, such as pathological image diagnosis~\cite{shao2021transmil, javed2022additive, qian2022transformer}: a whole slide image (WSI) is treated as a bag and a patch image cropped from WSI is treated as an instance.

Estimating instance labels from bag-level labels is studied as a weakly supervised learning. Many methods for this task use multi-label MIL (MLMIL), where multiple labels are attached to a bag, in which each label indicates whether a bag contains a corresponding class.
Mono-label Multi-class MIL is more challenging than MLMIL since only one label for each bag, i.e., a bag, is classified into a single class.
%\textcolor{red}{In addressing the multi-class MIL, where each bag is assigned a single class, there is a lack of research that organizes the relationship between bags and instances. }
The discussion about the relationship between bags and instances is insufficient in research on mono-label multi-class MIL.
If a bag contains instances of different classes, it is not uniquely determined which class should be assigned to the bag. %Many assumptions can be considered for multi-class MIL.

\begin{figure}[t]
 \begin{center}
    \includegraphics[width=0.58\columnwidth]{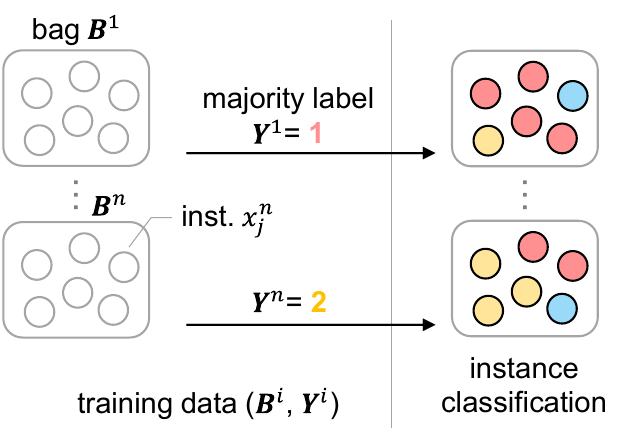}
 \vspace{-3mm}
    \caption{A novel problem setting: Learning from Majority.}
    \label{fig: intro}
 \end{center}
  \vspace{-7mm}
\end{figure}

This paper proposes a novel machine-learning problem in mono-label multi-class MIL called learning from the majority label (LML).
%\textcolor{red}{LML is a essential yet crucial problem where bag labels are determined by instance counts.}
LML aims to train a classification model to estimate the class of each instance using the majority label (Fig. \ref{fig: intro}). This problem setup is useful for many applications, such as medical image diagnosis. For example, in pathological image diagnosis, a majority class of cancer subtypes in a tissue is reported as diagnosis. 

% \label{sec:intro}
% \begin{figure}[t]
%  \begin{center}
%     \includegraphics[width=0.7\columnwidth]{./figs/intro.pdf}
%     \caption{Learning from Majority}
%     \label{fig: intro}
%  \end{center}
%  \vspace{-7mm}
% \end{figure}

\label{sec:intro}

% \begin{figure*}[t]
%  \begin{center}
%     \includegraphics[width=0.7\columnwidth]{./figs/counting_approach.pdf}
%     \caption{Counting approach}
%     \label{fig:counting}
%  \end{center}
%  \vspace{-7mm}
% \end{figure*}

% \begin{figure*}[t]
%  \begin{center}
%     \includegraphics[width=1.\columnwidth]{./figs/fig_method.pdf}
%     \caption{Overview of proposed method}
%     \label{fig: overview}
%  \end{center}
%   \vspace{-7mm}
% \end{figure*}

\begin{figure*}[htbp]
  \begin{minipage}[t]{0.28\linewidth}
    \centering
    \includegraphics[width=.84\columnwidth]{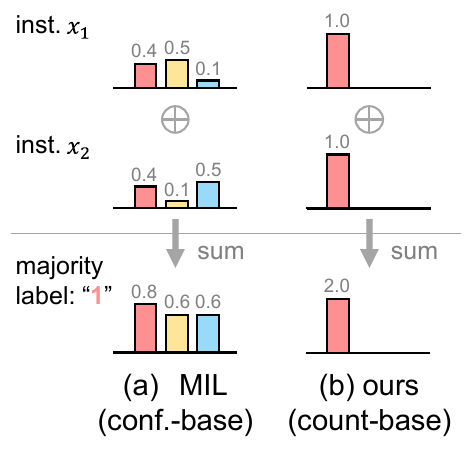}
    \vspace{-3mm}
    \caption{Counting approach.}
    \label{fig:counting}
  \end{minipage}
  \begin{minipage}[t]{0.65\linewidth}
    \centering
    \includegraphics[width=.86\columnwidth]{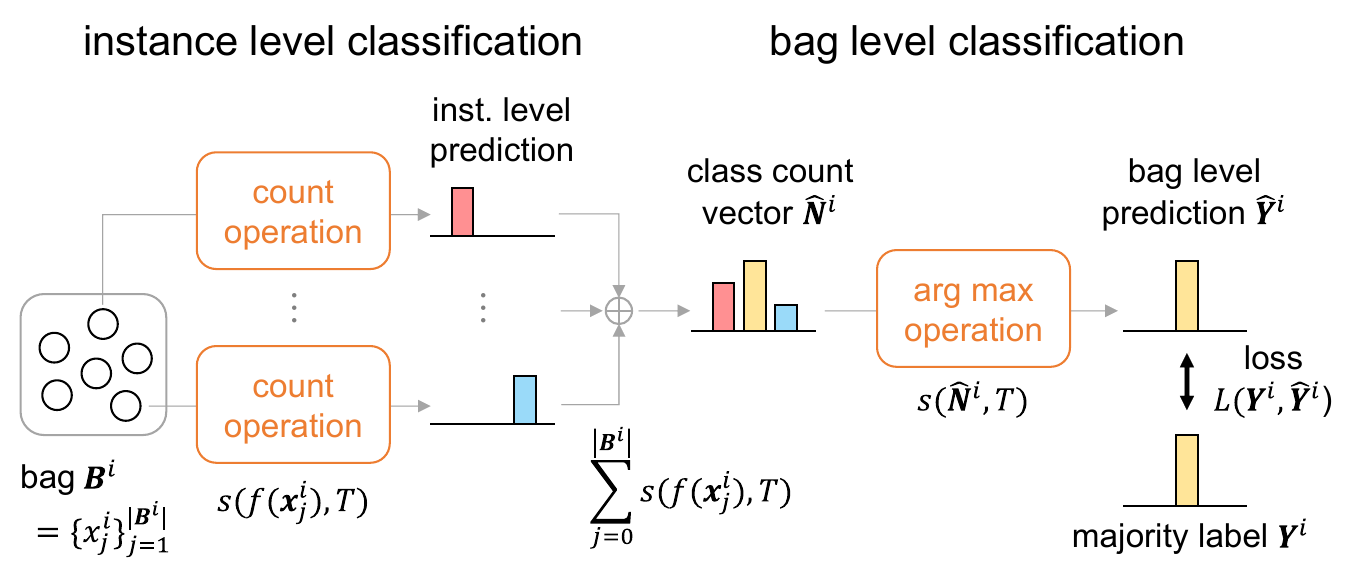}
    \vspace{-3mm}
    \caption{Overview of the proposed method.
    %, consisting of instance-level and bag-level classification by counting instance labels for each class.
    }
    \label{fig:overview}
  \end{minipage}
  \vspace{-3mm}
\end{figure*}

Existing MIL methods face challenges in solving LML.
For instance, the most popular MIL method predicts bag labels by summing the confidences of instances within a bag.
Let us consider a case when a bag contains two instances, and the majority label is class 1. In this case, the above MIL method can take various solutions that the maximum class of the sum of confidence becomes the majority class, e.g., (0.4, 0.5, 0.1) + (0.4, 0.1,0.5) = (0.8, 0.6, 0.6) or (0.4, 0.1, 0.5) + (0.6, 0.4, 0) = (1, 0.5, 0.5) as shown in Fig~\ref{fig:counting}.
It leads to the ambiguity of solutions and makes the problem difficult.
In addition, for these examples, the results of counting instances for each class,  (0,1,0)+(0,0,1)=(0,1,1) or (0,0,1)+(1,0,0)=(1,0,1), are inconsistent with the results of summing confidences.
This ambiguity and inconsistency lead training to fall to a bad local minimum. It decreases the accuracy of instance-level classification.

To solve these problems, we propose a counting network that counts the estimated class of instances in a bag to estimate the majority class. 
%\textcolor{red}{While our method is simple, the underlying discussion is non-trivial. }
By introducing a softmax with temperature, we can almost constrain confidences to binary values, in which the confidence of a class is close to 1 and the others close to 0 by the constraint. 
This enables the network to perform a counting operation with differentiable. This binary constraint for estimation avoids bad local minima in training.
Let us consider the above example; the majority label is class 1, and the bag size is 2. If the confidence only takes binary values, the solution is only one case: (1,0,0) + (1,0,0) = (2,0,0).
%\textcolor{red}{The proposed method achieved the highest accuracy in the experiments with four datasets compared to the conventional MIL. 
%In ablation studies, We demonstrated both logically and experimentally, the inconsistencies between the bag label by summing confidences and those from counting instances, decreasing accuracy. Our simple method improved the label consistency, significantly improving accuracy. These findings are significant progress in handling mono-label multi-classification MIL.}
The proposed method achieved the highest accuracy in the experiments with four datasets. Ablation studies were also conducted on our counting network, demonstrating its effectiveness.

The main contributions of this paper are summarized as follows:
\begin{itemize}
      \item We propose a novel problem, ``Learning from the Majority Label (LML),'' an essential yet crucial problem in mono-label multi-class MIL.
      \item We introduce a counting network for instance-level classification on LML because instance counts determine bag labels.
      \item We demonstrated, both logically and experimentally, the inconsistencies between the bag label by summing confidences and those from counting instances, decreasing accuracy, and our simple method improved the label consistency, significantly improving accuracy.
\end{itemize} 

\section{Related Work}
%\subsection{Multiple Instance Learning (MIL)}
\noindent
{\bf Multiple Instance Learning (MIL):}
In MIL, a bag consists of a set of instances and the class of a bag is given as supervised data, but that of each instance is unknown. MIL aims to train a classification model using bag-level labels.
Many MIL methods take an aggregation approach from instances in a bag, classified into output aggregation and feature aggregation.
Output aggregation~\cite{ramon2000multi, wang2018revisiting} aggregates the confidences of each instance estimated by a classification network, where the aggregated confidence is used as the bag-level class confidence and the network is trained by the classification loss, such as cross-entropy of the estimated bag-level label and its ground truth.
Feature aggregation~\cite{Ilse2020DeepMI, Pinheiro2015,ilse2018attention, shao2021transmil, rymarczyk2021kernel} aggregates the features of each instance extracted by a feature extractor and then inputs the aggregated features to a classifier to estimate the bag-level label. This approach has shown progressive results.
For the aggregation approach, many methods have been proposed: aggregating outputs or features by mean~\cite{wang2018revisiting, Ilse2020DeepMI}, max~\cite{Ilse2020DeepMI}, P-norm, and log-sum-exponential (LSE)~\cite{Pinheiro2015} operation. Recent methods use attention for aggregation, which takes the summation of weighted features by classical attention mechanism~\cite{ilse2018attention, javed2022additive, shi2020loss} or self-attention~\cite{shao2021transmil, javed2022additive, rymarczyk2021kernel}. 
The majority label has not been handled in the previous works.
In addition, conventional MIL methods focus on bag-level classification but not on instance-level.

%\subsection{Instance classification in MIL}
\noindent
{\bf Instance classification in MIL:}
Instance classification using bag-level labels can be considered weakly supervised learning, where a bag-level label is a weak label for the task of instance label estimation.
As applications of weakly supervised learning, image segmentation from class labels has been widely studied~\cite{Pinheiro2015,Oquab2015,kolesnikov2016seed,huang2018dsrg,Wang_2020_CVPR_SEAM}. In this task, a pixel or patch image is treated as an instance, and an image is treated as a bag, which is a set of pixels or patches. For example, binary classification MIL is used to segment tumor regions from non-tumor regions in a pathological image using image-level (bag-level) class labels~\cite{qian2022transformer}. 
For multi-class classification tasks, multi-label MIL (MLMIL) is used~\cite{zhou2012multi,yang2017miml,feng2017deep}. In this task, multiple labels are attached to a bag, where the number of labels is the same as that of classes, and each indicates whether a bag contains a corresponding class, i.e., binary classification for each class. Our task LML is much more challenging than MLMIL since a bag may contain different classes, and we only know the majority class and do not know whether the bag contains a different class or not.

\section{Learning from Majority}
\vspace{-2mm}
\subsection{Problem setting of Learning from Majority}
\vspace{-2mm}
LML is a kind of multiple instance learning (MIL).
LML aims to train a model for estimating a class of each instance using bag-level majority labels.
The details of the problem setup of LML are described as follows.

In LML, the set of bags and their majority labels $\mathcal{B} = \{ \bm{{B}}^i, \bm{Y}^i \}_{i=1}^n$ are given as training data, where each bag $\bm{B}^i$ consists of a set of instances, $\bm{B}^i = \{\bm{x}_j^i\}_{j=1}^{|\bm{B}^i|}$. $|\bm{B}^i|$ is the number of instances in $\bm{B}^i$.
$\bm{Y}^i \in \{0,1\}^C$ is an one-hot vector that represent majority class of $\bm{B}^i$, where $C$ is bagg..the number of class.

We here explain the relationship between a bag-level label $\bm{Y}^i$ and the instance-level labels of the bag $\bm{y}_j^i \in \{0,1\}^C$, $j=1 ,..., |\bm{B}^i|$. Note that the instance-level labels are unknown in training.
The number of instances that belongs to class $k$ in $\bm{B}^i$ is defined as:  
\begin{equation}
\label{eq:countVec}
N^i_k=\sum_{\bm{y}_j^i \in \bm{B}^i}{y_{j,k}^i}, \\
\end{equation}
where $y_{j,k}^i$ is the $k$-th element (class $k$) of $\bm{y}_{j}^i$. 
The summation indicates the counting operation of instances.
Counting vector $\bm{N}^i$ is defined as $\bm{N}^i=(N_1^i ,\ldots, N_k^i ,\ldots, N_C^i)^T$, which shows the count results for each class.

The majority class can be obtained by $\arg\max$ operation that finds the index whose value is maximum in $\bm{N}^i$.
The majority label $\bm{Y}^i$ is obtained as follows:

\begin{equation}
\label{eq:majorityLabel}
Y_c^i = \left\{
\begin{array}{ll}
1, & \mbox{if}\hspace{2mm} c =  \underset{{k =1,..,C}}{\arg \max} {N_k^i}\\
%\mbox{if}\hspace{2mm} c =  \underset{{k =1,..,C}}{\arg \max} \sum_{y_j^i \in \bm{B}^i}{y_{j,k}^i}, \\
0, & \mbox{otherwise},
\end{array}
\right. \\
\end{equation} 
where $Y_c^i$ is corresponding to class $c$.

In LML, we train a model to predict the labels of instances using only the bag-level labeled data $\mathcal{B}$. 
%This problem setup is a kind of weakly supervised learning: 
It can be considered as an inverse problem of the above counting operation.
This inverse problem is ill-posed, even in training, since there are many solutions of instance labels; any conditions satisfy the majority label when the number of instances belonging to the majority class is larger than those in the other classes.

\subsection{Counting network for learning from majority label}

Based on the problem setup, we proposed a counting network for LML, which is trained so that the estimated majority label by counting the class labels of instances should be the same as the given bag-level majority label.
Fig.~\ref{fig:overview} shows an overview of the proposed method, which consists of two steps: instance-level classification with a counting operation and bag-level classification, which obtains the majority class based on the counting results.

%%ここの部分は方法の概要を述べている感じでしょうか？
Standard classification methods estimate the class confidences, in which the sum of confidences is 1 by soft-max operation, and it takes a real value from 0 to 1.
As we introduced in Sec.1, the sum of confidences may become inconsistent with the number of instances.

In contrast, our method counts the number of instances with each class.
The count operation is realized by making the outputs of instance-level classification close to 1 or 0.
Consequently, the sum of outputs is consistent with the number of instances. 
Moreover, the network outputs the estimated class with the largest number of instances by using $\arg\max$ operation, where the standard network outputs the confidence value.
By training the network with a cross-entropy loss between the estimated class and the ground-truth majority label, we can obtain a network that outputs the bag-level classification by counting the number of instances for each class. 

%\subsubsection{Instance-Level classification}
\noindent
{\bf Instance-Level classification:} 
Given a bag $\bm{B}^i = \{\bm{x}_j^i\}_{j=1}^{|\bm{B}^i|}$, instance-level classifier $g: \mathbb{R}^{w \times h \times d}\rightarrow [0,1]^{C}$ estimates the class of each instance.
The classifier $g$ consists of network $f : \mathbb{R}^{w \times h \times d}\rightarrow \mathbb{R}^{C}$ and softmax function with temperature $s: \mathbb{R}^{C}\rightarrow [0,1]^{C}$.
Instead of employing the conventional softmax function, we employed a softmax with temperature to execute a counting operation. Softmax with temperature can control the distribution of the network output with differentiable. 
Using a small temperature ensures that the estimated value closely approximates either 0 or 1, where if an instance belongs to the class, it takes 1; otherwise, 0. 
Given a vector $\bm{z} \in \mathbb{R}^C$, this function outputs the normalized vector as follows:
\begin{equation}
\label{eq:softmax1}
\begin{split}
&s(\bm{z},T) = (s(\bm{z},T)_1,...,s(\bm{z},T)_c,...,s(\bm{z},T)_C), \\
&s(\bm{z},T)_c=\frac{\exp(\frac{z_{c}}{T})}{\sum\limits_{k=1}^C \exp(\frac{z_{k}}{T})},
\end{split}
\end{equation}
where $s(\bm{z},T)_c$ is the $c$-th element (class $c$) of output vector $s(\bm{z},T)$.
When the temperature $T$ is set to a low value, the network output represents a pseudo-one-hot vector, i.e., only a single class value approximately takes 1, and the others take 0.

%\subsubsection{Bag-Lebel classification by counting instance label}
\noindent
{\bf Bag-Level classification by counting instance label:} 
Next, given a set of estimation results in a bag, $\{g(\bm{x}_j^i,T)\}_{j=1}^{|\bm{B}^i|}$, this step counts instances in each class by a summation operation for the estimated pseudo-one-hot vector as follows:
\begin{equation}
\label{eq:sum}
\hat{N^i_c}=\sum\limits_{j=1}^{|B^i|} g(\bm{x}_j^i,T)_c,
\end{equation}
where $\hat{N^i_c}$ shows the number of instances that belong to class $c$ since $g(\bm{x}_j^i,T)_c$ approximately takes 0 or 1.

The majority class in the bag is the class that $\hat{N^i_c}$ is the largest among all classes, which can be obtained by $\arg\max$ operation ($\arg\max_{c}{\hat{N^i_c}}$). 
Suppose we use the confidence scores derived from the standard softmax function for the $\arg\max$ operation. In that case, the loss continues to decrease until all instances are classified as belonging to the majority class, even after the predicted class for a bag becomes the majority class; i.e., the standard softmax can lead to overestimating the majority class.
In order to avoid this overestimation, we also use softmax with temperature $s(\bm{N}^i, T)$ to take a majority class with a differentiable process. This process outputs the one-hot vector $\hat{\bm{Y}}^i=s(\bm{N}^i, T)$, where each element of $\hat{\bm{Y}}$ approximately takes 0 or 1, and if 1, the majority class of the bag.
Thus, the loss does not give a penalty when the estimated class of the bag is the same as the ground truth.

To train the network $g$, we use a cross-entropy between the estimated majority class $\hat{\bm{Y}}^i$
and its ground truth $\bm{Y}^i$ for each bag as:
\begin{equation}
\label{eq:loss}
L(\bm{Y}^i, \hat{\bm{Y}}^i) = -\sum\limits_{c=1}^C {Y}_{c}^i \log \hat{Y_{c}^i},
\end{equation}
where this loss is calculated for every bag in the bag-level supervised data $\mathcal{B} = \{ \bm{{B^i}}, \bm{Y}^i \}_{i=1}^n$.
We expect that the network obtains representation ability to classify each instance by this training process using bag-level supervised data for many bags.
In inference, the majority class of an instance is obtained by the classifier $g$.

\begin{table*}[t]
    \def\@captype{table}
      \makeatother
        \centering
        \caption{Instance-level accuracies of the comparative methods on four datasets.}
        \scalebox{0.8}[0.8]{
        \begin{tabular}{c||cccc|c||cccc|c||cccc|c} 
        \hline
        & \multicolumn{5}{c||}{Small} & \multicolumn{5}{c||}{Various} & \multicolumn{5}{c}{Large}\\
        \cline{2-16}
        Method &  CIFAR & SVHN & PATH & OCT & Average &  CIFAR & SVHN & PATH & OCT & Average &   CIFAR & SVHN & PATH & OCT & Average\\ \hline \hline
        Output+Mean~\cite{wang2018revisiting}  & 0.178 & 0.172 & 0.284 & 
0.279 & 0.228 & 0.371 & 0.528  & 0.660  & 0.552 & 0.422  & 0.532 & 0.808 & 0.732 & 0.675 & 0.687 \\
        Feature+Mean~\cite{Ilse2020DeepMI} & 0.297 & 0.395 & 0.462 & 0.265 &
        0.355 & 0.400  & 0.552  & 0.582  & 0.538 & 0.518 & 0.460  & 0.667 & 0.616 & 0.656 & 0.600\\
        Feature+Max~\cite{Ilse2020DeepMI} & 0.178 &
 0.108  & 0.367 & 0.256 & 0.227 & 0.293  & 0.202  & 0.508  & 0.337 & 0.335 & 0.340 & 0.276 & 0.512 & 0.370 & 0.375\\
        Feature+P-norm~\cite{Ilse2020DeepMI} & 0.268 &  0.172 & 0.487 & 0.287 & 0.304 & 0.351  &  0.289  & 0.578  & 0.448 & 0.417 & 0.402 & 0.399 & 0.568 & 
0.490 & 0.465\\
        Feature+LSE~\cite{Pinheiro2015} & 0.204 & 0.120 & 0.374 & 0.268 & 0.242 & 0.294 & 0.180  & 0.469  & 0.331 & 0.319 & 0.341 & 0.229 & 0.518 & 0.346 & 0.359\\
        Feature+Attention~\cite{ilse2018attention} & 0.163 & 0.108 & 0.350 & 0.263 & 
0.221 & 0.369 & 0.353  & 0.514  & 0.494 & 0.433 & 
0.406 & 0.500 & 0.591 & 0.561 & 0.515\\
        AdditiveMIL~\cite{javed2022additive} & 0.147 & 0.110 & 0.254 & 0.255 & 0.192 & 0.247  & 0.247  & 0.475  & 0.334 & 0.326 & 0.406 & 0.524 & 0.599 & 0.411 & 0.485\\
        AdditiveTransMIL~\cite{javed2022additive} & 0.270 & 0.154 & 
0.488 & 0.306 & 0.305 & 0.346  & 0.306  & 0.555  & 0.463 & 0.418 & 0.390 & 0.478 & 0.520 & 0.536& 0.481\\ 
        Ours   &\textbf{0.335} & \textbf{0.479} &   \textbf{0.578} & \textbf{0.392} & \textbf{0.446} &  \textbf{0.487}  & \textbf{0.721}  & \textbf{0.675}  & \textbf{0.627} & \textbf{0.628} &
        \textbf{0.614}  & \textbf{0.847} & \textbf{0.805} & \textbf{0.699} & \textbf{0.741}\\
        \hline
        \end{tabular}
        }
        \label{tab:comparison}
\vspace{-6mm}
\end{table*}

\begin{table*}[t]
    \def\@captype{table}
      \makeatother
        \centering
        \caption{Ablation study. Instance-level accuracies of the methods that use each operation on four datasets.}
        \scalebox{0.75}[0.75]{
        \begin{tabular}{c|cc||cccc|c||cccc|c||cccc|c} 
        \hline
        && Max & \multicolumn{5}{c||}{Small} & \multicolumn{5}{c||}{Various} & \multicolumn{5}{c}{Large}\\  \cline{4-18}
        Method & Count & class  &  CIFAR & SVHN & PATH & OCT & Average &  CIFAR & SVHN & PATH & OCT & Average &  CIFAR & SVHN & PATH & OCT & Average\\ \hline \hline
        Output+Mean~\cite{wang2018revisiting}  &&& 0.178 & 0.172 & 0.284 & 
0.279 & 0.228 & 0.371 & 0.528  & 0.660  & 0.552 & 0.422  & 0.532 & 0.808 & 0.732 & 0.675 & 0.687 \\
        Ours w/o Count && $\checkmark$ & 0.318 & 0.390 & 0.535 & 0.265  & 0.377  & 0.467  & 0.718  & 0.655  & 0.593 & 0.608 & 0.589 & \textbf{0.852} & 0.782 & \textbf{0.719} & 0.736\\
        % Ours w/o Max class &$\checkmark$&& 0.415 & 0.663 & 0.646 & \textbf{0.660} \\
        Ours   &$\checkmark$& $\checkmark$ & \textbf{0.335} & \textbf{0.479} &   \textbf{0.578} & \textbf{0.392} & \textbf{0.446} &  \textbf{0.487}  & \textbf{0.721}  & \textbf{0.675}  & \textbf{0.627} & \textbf{0.628} &
        \textbf{0.614}  & 0.847 & \textbf{0.805} & 0.699 & \textbf{0.741} \\ 
        \hline
        \end{tabular}
        }
        \label{tab:ablation}
        \vspace{-3mm}
\end{table*}

\section{Experiments}
%\subsection{Datasets and experimental setting}
\noindent
{\bf Datasets and experimental setting:} 
We used four datasets in our experiments. 1) CIFAR10 \cite{krizhevsky2009learning},  and 2) SVHN \cite{netzer2011reading} , these are commonly used in classification task. 
3) PATHMNIST and 4) OCTMNIST, these are datasets in MedMNIST \cite{yang2023medmnist}. Here, we choose large datasets in MedMNIST to prepare enough bags for this task.

The proportion of the majority class, which is the ratio of the number of instances of the majority class over the number of all instances in a bag, influences the difficulty of this task (LML).
For example, when the proportion is large, almost all instances in the bag belong to the majority class, making training easy. In contrast, when it is small, there are instances with various classes in a bag, and the number of instances with the first and the second majority classes is not so different, making training difficult.
Therefore, we conducted three scenarios: 1) Large when the majority proportion takes a large value (0.6 to 1); 2) Small when the majority proportion takes a small value (1/$C$ to 0.4), where $C$ is the number of classes; 3) Various when the majority proportion takes various values (1/$C$ to 1). In each scenario, the proportions of the classes were randomly determined, and instances were randomly sampled along with the proportions to make a bag.

%\subsection{Implementaion detail and metric}
\noindent
{\bf Implementation detail and performance metric:} 
We implemented our method by using PyTorch \cite{Paszke2019PyTorchAI}.
The network model $f$ was ResNet18 \cite{he2016deep} initialized by random weights. We used Adam Optimizer \cite{adam} with a learning rate of 3e-4 and epoch = 1500 to train our network.
The performance of each method is evaluated at the epoch with the best validation loss.
We set the mini-batch size as 64 and temperature $T$ = 0.1.
We evaluated our model performance using instance-level classification accuracy.

%\subsection{Evaluation with comparative methods}
\noindent
{\bf Evaluation with comparative methods:} 
To show the difficulty of LML for the current MIL methods and the effectiveness of our method, we compared our method with eight current methods, including state-of-the-art methods: 1) ``Output+Mean''\cite{ wang2018revisiting}, which aggregates the estimated class confidences from all instances in a bag and computes the loss using the aggregated vector; and 2) to 6) ``Feature+$\alpha$'', which aggregates the extracted features from all instances in a bag; currently, this method is major than output aggregation, and thus, we compared many types of aggregation methods. For aggregation ($\alpha$), we used various types of methods: 2) Mean~\cite{Ilse2020DeepMI}; 3) Max~\cite{Ilse2020DeepMI}; 4) P-norm; 5) LSE~\cite{Pinheiro2015}; 6) Attention~\cite{ilse2018attention}; 7) AdditiveMIL~\cite{javed2022additive}, and 8) AdditiveTransMIL~\cite{javed2022additive}, which are the state-of-the-art methods for standard MIL tasks.

\begin{figure}[th]
 \begin{center}
\includegraphics[width=0.65\columnwidth]{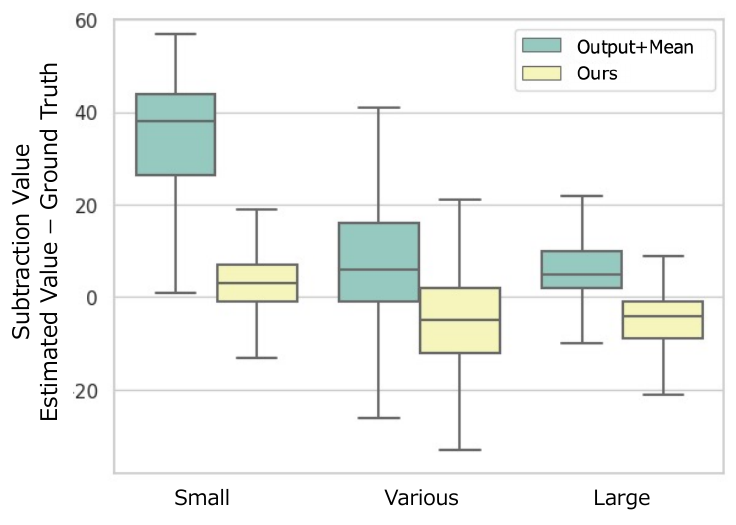}
    \caption{Effect of $\arg\max$ operation in each scenario. The subtraction value is average on all datasets.}
    \label{fig:MajorityProportion}
 \end{center}
 \vspace{-9mm}
\end{figure}

Table~\ref{tab:comparison} shows the instance-level accuracies of the comparative methods in three scenarios about the majority proportion.
Current MIL methods did not work for LML in all datasets and scenarios. These methods tend to overfit to the majority class even in training.
In addition, these methods do not consider the relationship between the instance labels and the bag label in the LML setting.
% \textcolor{red}{The Additive MIL addresses multi-class classification as multi-label MIL. This implies more cues are given as labels.}
Therefore, the accuracy was not good for test data.
Our method outperformed these comparative methods.
Comparing the scenarios, as we expected, the accuracies in ``Large'' were better than in ``Various'' and ``Small''. The scenario ``Small'' is challenging, and thus, the accuracies of all methods significantly decreased.

%\subsection{Ablation study}
\noindent
{\bf Ablation study:} 
An ablation study demonstrates the effectiveness of the counting operation and $\arg\max$ operation. ``Output+Mean'' is the baseline method, aggregating the confidence of instances in a bag by mean operation for bag-level estimation. 
``Ours w/o Count'' used the standard softmax for instance-level classification instead of softmax with temperature and represented the bag label by the sum of confidence of all instances in a bag.

Table~\ref{tab:ablation} shows the accuracies of the baseline methods. This result shows both operations improved the accuracies in all scenarios.
Without using the $\arg\max$ operation, as discussed above, the baseline method overestimated the majority class of the bag. This operation improved the performance of all datasets.

To verify the overestimation of the baseline method, we evaluated the difference in the number of instances of the majority class between the estimated one and the ground truth.
%proportion between the estimated one and the ground truth. 
Figure~\ref{fig:MajorityProportion} shows the box plot in each scenario, where the vertical axis indicates the subtraction value (estimated value - ground truth). If the number of instances is the same as the ground truth, it takes 0. The subtraction value above 0 indicates overestimation. The results show that the baseline method fell overestimation in all classes.
In particular, the overestimation of the baseline is significant in the scenario ``Small''.

% \begin{figure}[th]
%  \begin{center}
% \includegraphics[width=0.7\columnwidth]{./figs/effectiveness_max_class.pdf}
%     \caption{Distance between GT and predict instance number}
%     \label{fig:MajorityProportion}
%  \end{center}
%  \vspace{-7mm}
% \end{figure}

\begin{table}[t]
    \def\@captype{table}
      \makeatother
        \centering
        \caption{Consistency rate between labels estimated by aggregation and by counting instances. The rate is the average on all datasets.}
        \scalebox{0.8}[0.8]{
        \begin{tabular}{c|c|c|c} 
         \hline
          Method & \multicolumn{1}{c|}{Small} & \multicolumn{1}{c|}{Various} & \multicolumn{1}{c}{Large}\\ \hline 
        Output+Mean~\cite{wang2018revisiting}  & 0.957 & 0.961 & 0.997\\
        Feature+Mean~\cite{Ilse2020DeepMI}  & 0.690 & 0.892 & 0.987\\
        Feature+Max~\cite{Ilse2020DeepMI}   & 0.523 & 0.456 & 0.568\\
        Feature+P-norm~\cite{Ilse2020DeepMI}  & 0.458 & 0.762 & 0.925\\
        Feature+LSE~\cite{Pinheiro2015}   & 0.290 & 0.494 & 0.609\\
        Feature+Attention~\cite{ilse2018attention}  & 0.831 & 0.730 & 0.911\\
        AdditiveMIL~\cite{javed2022additive}   & 0.526 & 0.469 & 0.736\\
        AdditiveTransMIL~\cite{javed2022additive}   & 0.494 & 0.780 & 0.889\\
        Ours w/o Count  & 0.914 & 0.973 & 0.999\\
        Ours  & \textbf{0.963} & \textbf{0.990} & \textbf{1.00}\\
        \hline
        \end{tabular}
}
        \label{tab:consistency}
        \vspace{-3mm}
\end{table}

In addition, the counting operation further improved the accuracy. In particular, this greatly improved the accuracy in the scenario ``Small''. When the majority proportion is small, the standard confidence summation tends to fall to a local minimum since the problem becomes more ambiguous; the proportion of the second majority class is close to that of the first one.

%The advantage of the counting operation is the consistency of the majority class between one obtained by summing the network outputs and obtained by counting instances for each class.
Table~\ref{tab:consistency} shows the consistency rate when the estimated majority class was correct.
The consistency rate is defined as the number of cases when the majority class obtained by counting the instances is the same as that obtained by aggregating the network outputs divided by the number of all bags when the estimated majority class is correct.
The current feature aggregation MIL methods produced a different majority class between that obtained by counting instances and aggregation since the classifier inputs are different: one is a feature extracted from each instance, and the other is the aggregated feature. Therefore, their consistency rate is low.
Output+Mean aggregates the network outputs, and the classifier inputs for an instance and a bag are the same; the consistency rate was improved from feature aggregation.
However, these methods still lead to inconsistency with the instance count results. In contrast, the consistency rate of our method was better than that of the comparative methods because of our counting approach.
The improvement of our approach showed more significance in "Small." This difference arises from amplifying the chance of a label switching to the minor class in the baseline methods when ``Small'' because the difference in the sum of confidence between the major and minor classes becomes small.

%The baseline methods in the ablation study, which do not use the counting operation, aggregate the network outputs. So, the classifier inputs for an instance and a bag are the same; thus, the consistency rate was improved from feature aggregation.
%However, these methods use summation of confidences; it is not counting operation. 
%\textcolor{red}{Baseline method often leads to inconsistency with the instance count results when the majority proportion is small.  When large, the sum of the confidences for the minority class is small, reducing the chance of a label switching to the minor class. That’s why counting improved accuracies when small.}

\section{Conclusion}
This paper proposes a novel problem, ``Learning from 
the Majority Label (LML),'' which trains a model to classify each instance using the bag-level majority labels without instance-level labels. We show the relationship between the instance and bag labels; the bag-level label can be obtained by counting the instance labels.
The current MIL aggregating confidence is unsuitable for this task since confidence aggregation is different from counting. To solve LML, we propose a counting network that counts the estimated class of instances in a bag to estimate the majority class. In experiments, we show the effectiveness of our counting network, and our method outperformed current MIL methods in LML.

%\section{Acknowledgments}

% \noindent
% {\bf Acknowledgements}: This work was supported by JSPS KAKENHI Grant Number JP23K18509, and JST, ACT-X Grant Number JPMJAX200G, Japan. \\

% \vspace{90pt}
\bibliographystyle{IEEEbib}
\bibliography{refs}

\end{document}